\documentclass[runningheads]{llncs}
\usepackage{graphicx}
\usepackage{amsmath,amssymb} 
\usepackage{color}

\usepackage{subfig}
\usepackage{hyperref}

\begin{document}

\title{Summarizing Videos with Attention\thanks{This research was funded by the H2020 MONICA European project 732350 and by the NATO within the WITNESS project under grant agreement number G5437 and  within the MIDAS G5381. We gratefully acknowledge the support of NVIDIA Corporation with the donation of the Titan Xp GPU used for this research.}} 
\titlerunning{Summarizing Videos with Attention}


\author{Jiri Fajtl\inst{1} \and
	Hajar Sadeghi Sokeh\inst{1}\and
	Vasileios Argyriou\inst{1} \and
	Dorothy Monekosso\inst{2} \and
	Paolo Remagnino\inst{1}}
%

\authorrunning{J. Fajtl et al.} 


\institute{
Robot Vision Team RoVit, Kingston University, London, UK \and	
	Leeds Beckett University, Leeds, UK}

\maketitle


\begin{abstract}
In this work we propose a novel method for supervised, keyshots based video summarization by applying a conceptually simple and computationally efficient soft, self-attention mechanism. Current state of the art methods leverage bi-directional recurrent networks such as BiLSTM combined with attention. These networks are complex to implement and computationally demanding compared to fully connected networks. To that end we propose a simple, self-attention based network for video summarization which performs the entire sequence to sequence transformation in a single feed forward pass and single backward pass during training. Our method sets a new state of the art results on two benchmarks TvSum and SumMe, commonly used in this domain.

\end{abstract}

\section{Introduction}
Personal videos, video lectures, video diaries, video messages on social networks and videos in many other domains are becoming to dominate other forms of information exchange. According to Cisco Visual Networking Index: Forecast and Methodology, 2016-2021\footnote{\url{https://www.cisco.com/c/en/us/solutions/collateral/service-provider/visual-networking-index-vni/complete-white-paper-c11-481360.html}}, by 2019 video will account for 80\% of all global Internet traffic, excluding P2P channels. Consequently, better methods for video management, such as video summarization, are needed.

Video summarization is a task where a video sequence is reduced to a small number of still images called keyframes, sometimes called storyboard or thumbnails extraction, or a shorter video sequence composed of keyshots, also called video skim or dynamic summaries. The keyframes or keyshots need to convey most of key information contained in the original video. This task is similar to a lossy video compression, where the building block is a video frame. In this paper we focus solely on the keyshots based video summarization.

Video summarization is an inherently difficult task even for us people. In order to identify the most important segments one needs to view the entire video content and then make the selection, subject to the desired summary length.
Naturally, one could define the keyshots as segments that carry mutually diverse information while also being highly representative of the video source. There are methods that formulate the summarization task as a clustering with cost functions based on exactly these criteria. Unfortunately, to define how well chosen keyshots represent the video source as well as the diversity between them is extremely difficult since this needs to reflect the information level perceived by the user. Common techniques analyze motion features, measure the distance between color histograms, image entropy or in the 2/3D CNN feature space \cite{novak1992anatomy,DBLP:journals/corr/Larkin16,5508420,athiwaratkun2015feature}, reflecting semantic similarities. However, none of these approaches can truly capture the information in the video context. We believe that to automatically generate high quality summaries, similar to what we are capable of, a machine should learn from us humans by means of a behavioral cloning or supervision.

Early video summarization methods were based on unsupervised methods, leveraging low level spatio-temporal features and dimensionality reduction with clustering techniques. Success of these methods solely stands on the ability to define distance/cost functions between the keyshots/frames with respect to the original video. As discussed above, this is very difficult to achieve as well as it introduces a strong bias in the summarization given by the type of used features such as semantic and pixel intensities. 
In contrast, models trained with supervision learn the transformation that produces summaries similar to those manually produced. Currently, there are two datasets with such annotations, TvSum \cite{song2015tvsum} and SumMe \cite{GygliECCV14}, where each video is annotated by 15-20 users. The annotations vary between users with consistency expressed by a pairwise F-score $\sim$ 0.34. This fact reveals that the video annotation is a rather subjective task. We argue that under these circumstances it may be extremely difficult to craft a metric that would accurately express how to cluster video frames into keyshots, similar to human annotation. On this premise, we decided to adopt the supervised video summarization for our work.

\begin{figure*}[h]
\vspace{-15pt}
	\begin{center}
		\includegraphics[width=0.7\linewidth]{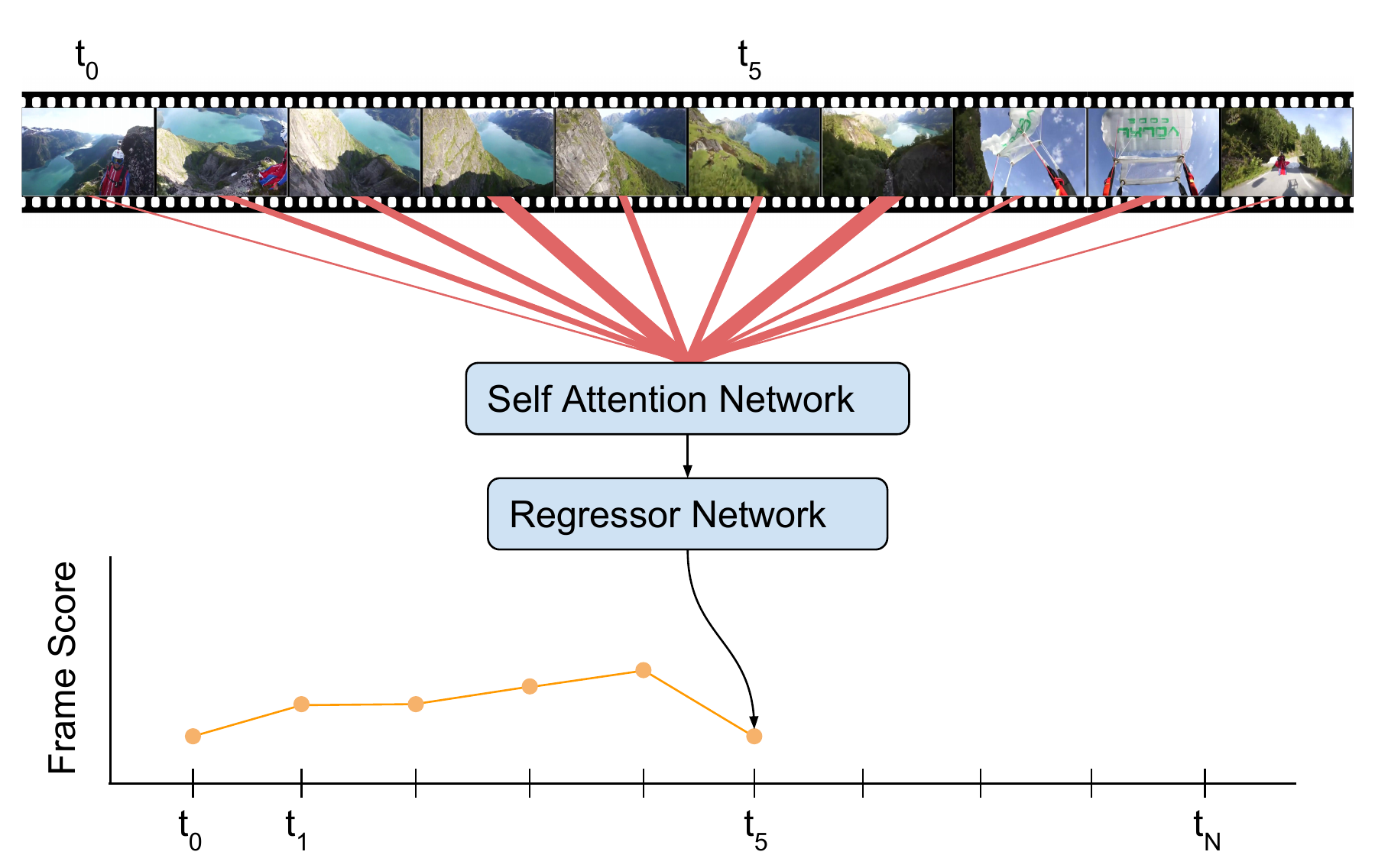}
	\end{center}
	\caption{For each output the self-attention network generates weights for all input features. Average of the input features, weighted by this attention, is regressed by a fully connected neural network to the frame importance score.}
\label{fig:patnet_high_level}
\end{figure*}

Current state of the art methods for video summarization are based on recurrent encoder-decoder architectures, usually with bi-directional LSTM \cite{hochreiter1997long} or GRU \cite{cho2014learning} and soft attention \cite{bahdanau2014_neural_machine_translation}. While these models are remarkably powerful in many domains, such as machine translation and image/video captioning, they are computationally demanding, especially in the bi-directional configuration. Recently A. Vaswani et al. \cite{NIPS2017_7181} demonstrated that it is possible to perform sequence to sequence transformation only with the attention. Along similar lines, we propose a pure attention, sequence to sequence network VASNet for video keyshots summarization and demonstrate its performance on TvSum and SumMe benchmarks. 
Architecture of this model does not employ recurrent or sequential processing and can be implemented with conventional matrix/vector operations and run in a single forward/backward pass during inference/training, even for sequences with variable length. The architecture is centered around two key operations, attention weights calculation and frame level score regression. An overview of this model is shown in Fig. \ref{fig:patnet_high_level}. 
Frame score at every step $t$ is estimated from a weighted average of all input features. The weights are calculated with the self-attention algorithm. Given the generic architecture of our model we believe that it could be successfully used in other domains requiring sequence to sequence transformation. Our contributions are:
\begin{enumerate}
	\item A novel approach to sequence to sequence transformation for video summarization based on soft, self-attention mechanism. In contrast, current state of the art relies on complex LSTM/GRU encoder-decoder methods.
	\item A demonstration that a recurrent network can be successfully replaced with simpler, attention mechanism for the video summarization. 
\end{enumerate}

\section{Related Work} 
Recent advancements in deep learning were rapidly adapted by researches focusing on video summarization, particularly encoder-decoder networks with attention for sequence to sequence transformation. In this section we will discuss several existing methods related to our work.

K. Zhang et al. \cite{zhang2016video} pioneered the application of LSTM for supervised video summarization to model the variable-range temporal dependency among video frames to derive both representative and compact video summaries. They enhance the strength of the LSTM with the determinantal point process which is a probabilistic model for diverse subset selection. 
Another sequence to sequence method for supervised video summarization was introduced by Ji et al. \cite{ji2017video}. Their deep attention-based framework uses a bi-directional LSTM to encode the contextual information among input video frames. 
Mahasseni et al. \cite{Mahasseni2017UnsupervisedVS} propose an adversarial network to summarize the video by minimizing the distance between the video and its summary. They predict video keyframes distribution with a sequential generative adversarial network. 
A deep summarization network in an encoder-decoder architecture via an end-to-end reinforcement learning has been proposed by Zhou et al. \cite{Zhou2018DeepRL} to achieve state of the art results in unsupervised video summarization. They design a novel reward function that jointly takes diversity and representativeness of generated summaries into account. 
A hierarchical LSTM is constructed to deal with the long temporal dependencies among video frames by \cite{zhao2017hierarchical}, but it fails to capture the video structure information, where the shots are generated by fixed length segmentation. 

Some works use side semantic information associated with a video along with visual features, like surrounding text such as titles, queries, descriptions, comments, unpaired training data and so on. 
Rochan et al. in \cite{rochan2018learning}, proposed deep learning video summaries from unpaired training data, which means they learn from available videos summaries without their corresponding raw input videos. Yuan et al. \cite{yuan2017video}, proposed a deep side semantic embedding model which uses both side semantic information and visual content in the video. Similarly H. Wei et al. \cite{DBLP:conf/aaai/WeiNYYYY18} propose a supervised, deep learning method trained with manually created text descriptions as ground truth. At the heart of this  method is the LSTM encode-decoder network. Wei achieves competitive results with this approach, however, more complex labels are required for the training. Fei et al. \cite{Fei:2017:MRV:3174277.3174287} complemented visual features with video frame memorability, predicted by a separate model such as \cite{khosla2015understanding} or \cite{fajtl2018amnet}.

Other approaches, like the one described in \cite{dos2016summarizing}, use an unsupervised method by clustering some features extracted from the video, delete the similar frames, and select the rest of the frames as keyframe of the video.  In fact, they used a hierarchical clustering method to generate a weight map from the frame similarity graph in which the clusters can easily be inferred.
Another clustering method is proposed by Otani et al. \cite{otani2016video}, in which they use deep video features to encode various levels of content including objects, actions, and scenes. They extract the deep features from each segment of the original video and apply a clustering-based summarization technique on them.

\subsection{Attention Techniques}
The fundamental concept of attention mechanism for neural networks was laid by Bahdanau et al. \cite{bahdanau2014_neural_machine_translation} for the task of machine translation. This attention is based on an idea that the neural network can learn how important various samples in a sequence, or image regions, are with respect to the desired output state. These importance values are defined as attention weights and are commonly estimated simultaneously with other model parameters trained for a specific objective. There are two main distinct attention algorithms, hard and soft. 

Hard attention produces a binary attention mask, thus making a 'hard' decision on which samples to consider. This technique was successfully used by K. Xu et al. \cite{show_attend_tell} for image caption generation. Hard attention models use stochastic sampling during the training; consequently, backpropagation cannot be employed due to the non-differentiable nature of the stochastic processes. REINFORCE 
learning rule \cite{williams1992simple} is regularly used to train such models. This task is similar to learning an attention policy introduced by V. Mnih et al. \cite{mnih2014recurrent}.

In this work we exclusively focus on soft attention. In contrast to the hard attention, soft attention generates weights as true probabilities. These weights are calculated in a deterministic fashion using a process that is differentiable. This means that we can use backpropagation and train the entire model end-to-end. Along with the LSTM, soft attention is currently employed in the majority of sequence to sequence models used in machine translation \cite{Luong2015EffectiveAT}, image/video caption generation \cite{show_attend_tell},\cite{yao2015describing}, addressing neural memory \cite{graves2016hybrid} and other. 
Soft attention weights are usually calculated as a function of the input features and the current encoder or decoder state. 
The attention is global if at each step $t$ all input features are considered or local where the attention has access to only limited number of local neighbors. 

If the attention model does not consider the decoder state, the model is called self-attention or intra-attention. In this case the attention reflects the relation of an input sample $t$ with respect to other input samples given the optimization objective. Self-attention models were successfully used in tasks such as reading comprehension, summarization and in general for task-independent sequence representations \cite{cheng2016long}\cite{D16-1244}\cite{lin+al-2017-embed-iclr}. The self-attention is easy and fast to calculate with matrix multiplication in a single pass for entire sequence since at each step we do not need the result of past state.

\section{Model Architecture}
Common approach to supervised video summarization and other sequence to sequence transformations, is an application of a LSTM or GRU encoder-decoder network with attention. Forward LSTM is usually replaced with bi-directional BiLSTM since keyshots in the summary have relation to future video frames in the sequence.
Unlike the RNN based networks, our method does not need to reach for special techniques, such as BiLSTM, to achieve non-causal behavior. The vanilla attention model has equal access to all past and future inputs. This aperture can be, however, easily modified and it can even be asymmetric, dilated, or exclude the current time step $t$.

The hidden state passed from encoder to decoder has always fixed length, however, it needs to encode information representing sequences with variable lengths. This means that there is a higher information loss for longer sequences. The proposed attention mechanism does not suffer from such loss since it accesses the input sequence directly without an intermediate embedding.
 
\begin{figure*}[h]
	\begin{center}
		\includegraphics[width=1.\linewidth]{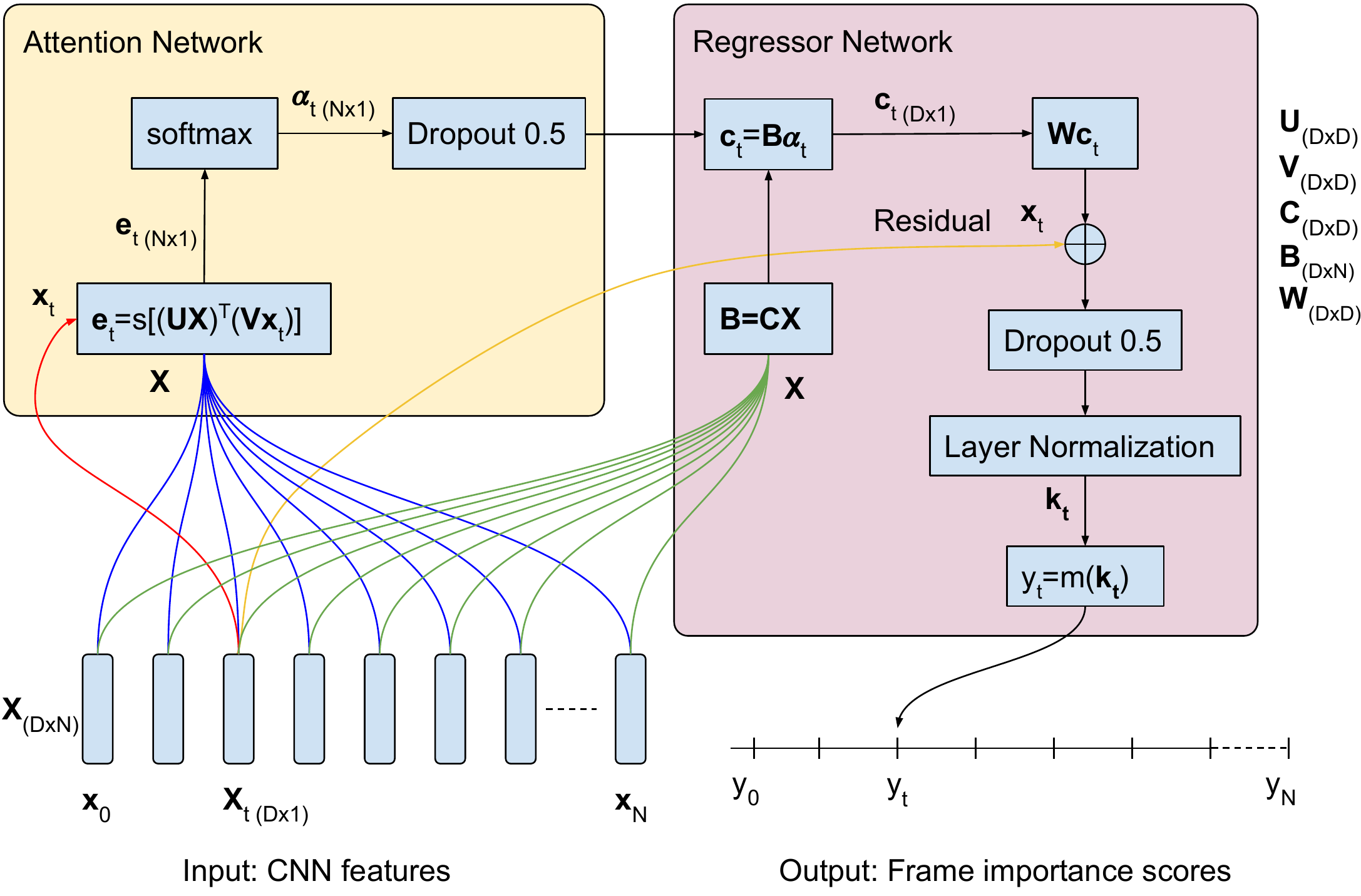}
	\end{center}
	\caption{Diagram of VASNet network attending sample $\boldsymbol{x}_t$.}
	\label{fig:patnet_detail}
\end{figure*}

Architecture proposed in this work replaces entirely the LSTM encoder-decoder network with the soft, self-attention and a two layer, fully connected network for regression of the frame importance score. 
Our model takes an input sequence $\boldsymbol{X}=(\boldsymbol{x}_0,\dots,\boldsymbol{x}_N), \quad \boldsymbol{x} \in \mathbb{R}^{D}$ and produces an output sequence $\boldsymbol{Y} = (y_0,\dots,y_N), \quad y=[0,1)$, both of length  $N$. The input is a sequences of CNN feature vectors with dimensions $D$, extracted for each video frame. Fig. \ref{fig:patnet_detail} shows the entire network in detail.

Unnormalized self-attention weight $e_{t,i}$ is calculated as an alignment between input feature $\mathbf{x}_t$ and the entire input sequence according to Luong et al. \cite{luong2015effective}. 
\begin{equation} \label{eq:e_dot}
e_{t,i} = s[ (\boldsymbol{U}\boldsymbol{x}_i)^T (\boldsymbol{V}\boldsymbol{x}_t)] \qquad t=[0,N), \quad i=[0,N)
\end{equation}
Here, $N$ is the number of video frames,  $\boldsymbol{U}$ and $\boldsymbol{V}$ are network weight matrices estimated together with other parameters of the network during optimization and $s$ is a scale parameter that reduces values of the dot product between $\boldsymbol{U}\boldsymbol{x}_i$ and $\boldsymbol{V}\boldsymbol{x}_t$. We set the scale $s$ to value 0.06, determined experimentally. Impact of the scale on the model performance was, however, minimal. 
Alternatively, the attention vector could be also realized by an additive function as shown by Bahdanou et al. \cite{bahdanau2014_neural_machine_translation}.
\begin{equation} \label{eq:e_sum}
e_{t,i} = \boldsymbol{M}\tanh(\boldsymbol{U}\boldsymbol{x}_i+\boldsymbol{V}\boldsymbol{x}_t)
\end{equation}
where $\boldsymbol{M}$ are additional network weights learned during training. 
Both formulas have shown similar performance, however, the multiplicative attention is easier to parallelise since it can be entirely implemented as a matrix multiplication which can be highly optimized.
The attention vector $\boldsymbol{e}_t$ is then converted to the attention weights $\boldsymbol{\alpha}_{t}$ with softmax.
\begin{equation} \label{eq:softmax}
{\alpha}_{t,i}={\frac {\exp({{e}_{t,i}})}{\sum _{k=1}^{N}\exp({{e}_{t,k}})}} \qquad 
\end{equation}
The attention weights $\boldsymbol{\alpha}_{t}$ are true probabilities representing the importance of input features with respect to the desired frame level score at the time $t$. 
Linear transformation $\boldsymbol{C}$ is then applied to each input and the results then weighted with attention vector $\boldsymbol{\alpha}_{t}$ and averaged. The output is a context vector $\boldsymbol{c}_t$ which is used for the final frame score regression.

\begin{equation} \label{eq:input_layer}
\boldsymbol{b}_i = \boldsymbol{C}\boldsymbol{x}_i
\end{equation}

\begin{equation} \label{eq:context_vector}
\boldsymbol{c}_t = \sum_{i=1}^{N} {\alpha}_{t,i} \boldsymbol{b}_i  \qquad \boldsymbol{c}_t \in \mathbb{R}^{D}
\end{equation}
The context vector $\boldsymbol{c}_t$ is then projected by a single layer, fully connected network with linear activation and residual sum 
followed by dropout and layer normalization. 
\begin{equation} \label{eq:y_regression}
\boldsymbol{k}_t = norm(dropout(\boldsymbol{W}\boldsymbol{c}_t + \boldsymbol{x}_t))
\end{equation}
The $\boldsymbol{C}$ and $\boldsymbol{W}$ are network weight matrices learned during the network training. To regularize the network we also add a dropout for attention weights as shown in Fig. \ref{fig:patnet_detail}. We found it to be beneficial, especially for small training datasets such as in the canonical setting for TvSum (40 videos) and SumMe (20 videos). 

By design, the attention network discards the temporal order in the sequence. This is due to the fact that the context vector $\boldsymbol{c}_t$ is calculated as a weighted average of input features without any order information. The order of the output sequence is still preserved. The positional order for the frame score prediction is not important in the video summarization task, as has been shown in the past work utilizing clustering techniques that also discard the input frame order.
For other tasks, such as  machine translation or captioning, the order is essential. In these cases every prediction at time $t$, including attention weights, could be conditioned on state at $t-1$. Alternatively, a positional encoding could be injected to the input as proposed by \cite{NIPS2017_7181},\cite{pmlr-v70-gehring17a}.

Finally, a two layer neural network  performs the frame score regression $y_t=m(\boldsymbol{k}_t)$. First layer has a ReLU activation followed by dropout and layer normalization \cite{ba2016layer}, while the second layer has a single hidden unit with sigmoid activation.

\subsection{Frame Scores to Keyshot Summaries}
\label{sec:scores_to_summaries}
\begin{figure*}[h]
	\begin{center}
		\includegraphics[width=0.8\linewidth]{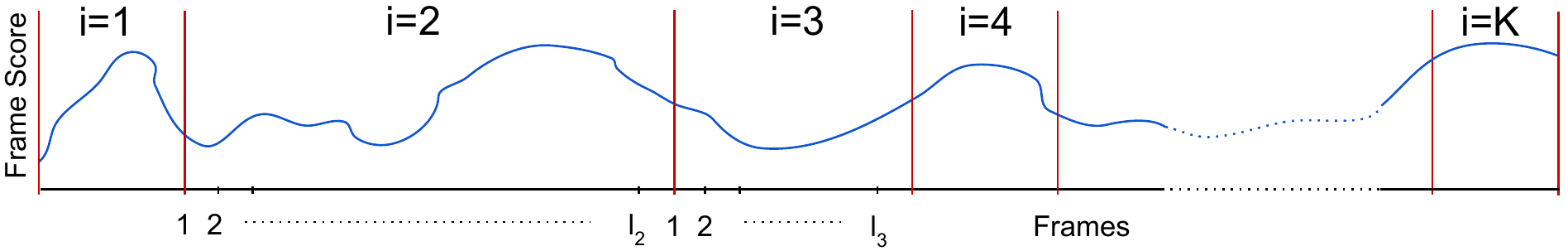}
		\vspace{-10pt}
	\end{center}
	\caption{Temporal segmentation with KTS.}
	\label{fig:kts_segmentation}
\end{figure*}

The model outputs frame-level scores that are then converted to keyshots. Following \cite{zhang2016video}, this is done in two steps. First, we detect scene change points where each represents a potential keyshot segment. Second, we select a subset of these keyshots by maximizing the total frame score within these keyshots while constraining the total summary length to 15\% of the original video length as per \cite{GygliECCV14}. The scene change points are detected by Kernel Temporal Segmentation (KTS) method \cite{potapov2014category} as shown in Fig. \ref{fig:kts_segmentation}.
For each detected shot $i\in K$ we calculate score $s_i$.
\begin{equation} \label{eq:e1}
s_i = \frac{1}{l_i}\sum_{a=1}^{l_i} y_{i,a}  
\end{equation}
where $y_{i,a}$ is score of $a$-th frame within shot $i$ and $l_i$ is the length of $i$-th shot.
Keyshots are then selected with the Knapsack algorithm 
Eq. \ref{eq:knapsack}
according to \cite{song2015tvsum}.
\begin{equation} \label{eq:knapsack}
\max \sum_{i=1}^K u_i s_i, \quad \text{s. t.} \quad \sum_{i=1}^K u_i l_i \le L, u_i \in {0,1}
\end{equation}
Keyshots with $u_i=1$ are then concatenated to produce the final video summary. For evaluation we create a binary summary vector where each frame in shot ($u_i=1$) is set to one. 

\subsection{Model Training}
To train our model we use the ADAM optimizer \cite{adam} with learning rate $5 \cdot 10^{-5}$ This low learning rate is used as a result of having a batch with single sample, where the sample is an entire video sequence. We use 50\% dropout and $L2=10^{-5}$ regularization. Training is done over 200 epochs. Model with the highest validation F-score is then selected.

\subsection{Computation Complexity}
The self-attention requires a constant number of operations at each step for all input features $N$, each of size $D$. The complexity is thus $O(N^2D)$. The recurrent layer, on the other hand, requires $O(N)$ sequential operations, each of complexity $O(N D^2)$. Self-attention needs less computation when the sequence length $N$ is shorter than the feature size $D$. For longer videos, a local attention would be used rather then the global one.

\section{Evaluation}

\subsection{Datasets Overview}
In order to directly compare our method with the previous work we conducted all experiments on four datasets, TvSum \cite{song2015tvsum}, SumMe \cite{GygliECCV14}, OVP \cite{de2011vsumm} and YouTube \cite{de2011vsumm}. OVP and YouTube were used only to augment the training dataset. TvSum and SumMe are currently the only datasets suitably labeled for keyshots video summarization, albeit still small for training deep models. Table \ref{table:datasets_properties} provides an overview of the main datasets properties.

\setlength{\tabcolsep}{4pt}
\begin{table}
	\begin{center}
		\caption{Overview of the TvSum and SumMe properties.}		
		\label{table:datasets_properties}

\begin{tabular}{l|c|l|l|c|c|c}
	\cline{1-7}
	\textbf{}& \textbf{}& & & \multicolumn{3}{l}{\textbf{Video length (sec)}} \\ 
	\cline{5-7} 
	\multicolumn{1}{l|}{\textbf{Dataset}} & \multicolumn{1}{c|}{\textbf{Videos}} & \multicolumn{1}{l|}{\textbf{\begin{tabular}[c]{@{}l@{}}User \\ annotations\end{tabular}}} & \textbf{\begin{tabular}[c]{@{}l@{}}Annotation\\ type\end{tabular}}      & \textbf{Min}   & \textbf{Max}   & \textbf{Avg}   \\ \hline
	\multicolumn{1}{l|}{\textbf{SumMe}}   & \multicolumn{1}{c|}{25}              & \multicolumn{1}{c|}{15-18}                                                                & keyshots                                                                & 32             & 324            & 146            \\ \hline
	\multicolumn{1}{l|}{\textbf{TvSum}}   & \multicolumn{1}{c|}{50}              & \multicolumn{1}{c|}{20}                                                                   & \begin{tabular}[c]{@{}l@{}}frame-level \\ importance scores\end{tabular} & 83             & 647            & 235            \\ \hline
	\multicolumn{1}{l|}{\textbf{OVP}}     & \multicolumn{1}{c|}{50}              & \multicolumn{1}{c|}{5}                                                                    & keyframes                                                                &  46              &  209              &   98             \\ \hline
	\multicolumn{1}{l|}{\textbf{YouTube}} & \multicolumn{1}{c|}{39}              & \multicolumn{1}{c|}{5}                                                                    & keyframes                                                                & 9               & 572               &  196              \\ \hline
\end{tabular}
	\end{center}
\end{table}

The TvSum dataset is annotated by frame-level importance scores, while 
the SumMe with binary keyshot summaries. OVP and YouTube are annotated with keyframes and need to be converted to the frame-level scores and binary keyshot summaries, following the protocol discussed in the following section \ref{sec:gt_preparation}.

\subsection{Ground Truth Preparation}
\label{sec:gt_preparation}
Our model is trained using frame-level scores, while the evaluation is performed with the binary keyshot summaries.
The SumMe dataset comes with keyshot annotations, as well as frame-level scores calculated as an average of the keyshot user summaries per frame. In the case of TvSum we convert the frame-level scores to keyshots following the protocol described in section \ref{sec:scores_to_summaries}. 
Keyframe annotations in OVP and YouTube are converted to frame-level scores 
by temporarily segmenting the video into shots with KTS and then selecting shots that contain the keyframes. Knapsack is then used to constrain the total summary length, however in this case the keyshot score $s_i$ (Eq. \ref{eq:knapsack}) is calculated as a ratio of number of keyframes within the keyshot and the keyshot length.

To make the comparison even more direct, we adopt identical training and testing ground truth data used by \cite{zhang2016video}, \cite{Zhou2018DeepRL} and \cite{Mahasseni2017UnsupervisedVS}. This represents CNN embeddings, scene change points, and generated frame-level scores and keyshot labels for all datasets.
The preprocessed data are publicly available (K. Zhou et al. \cite{Zhou2018DeepRL} \footnote{\url{http://www.eecs.qmul.ac.uk/~kz303/vsumm-reinforce/datasets.tar.gz}} and K Zhang et al.\cite{zhang2016video} \footnote{\url{https://www.dropbox.com/s/ynl4jsa2mxohs16/data.zip?dl=0}}). CNN embeddings used in this preprocessed dataset have 1024 dimensions and were extracted from the pool5 layer of the GoogLeNet network \cite{GoogLeNet} trained on ImageNet \cite{imagenet}.							

We use a 5-fold cross validation for both, canonical and augmented settings as suggested by \cite{zhang2016video}. In the canonical setting, we generate 5 random train/test splits for the TvSum and SumMe datasets individually. 80\% samples are used for training and the rest for testing. In the augmented setting we also maintain the 5-fold cross validation with the 80/20 train/test, but add the other datasets to the training split. 
For example, to train the SumMe in the augmented setting we take all samples from TvSum, OVP and YouTube and 80\% of the SumMe as the training dataset and the remaining 20\% for evaluation.

\subsection{Evaluation Protocol}
\begin{figure}[h]
	\begin{center}
		\includegraphics[width=1.\linewidth]{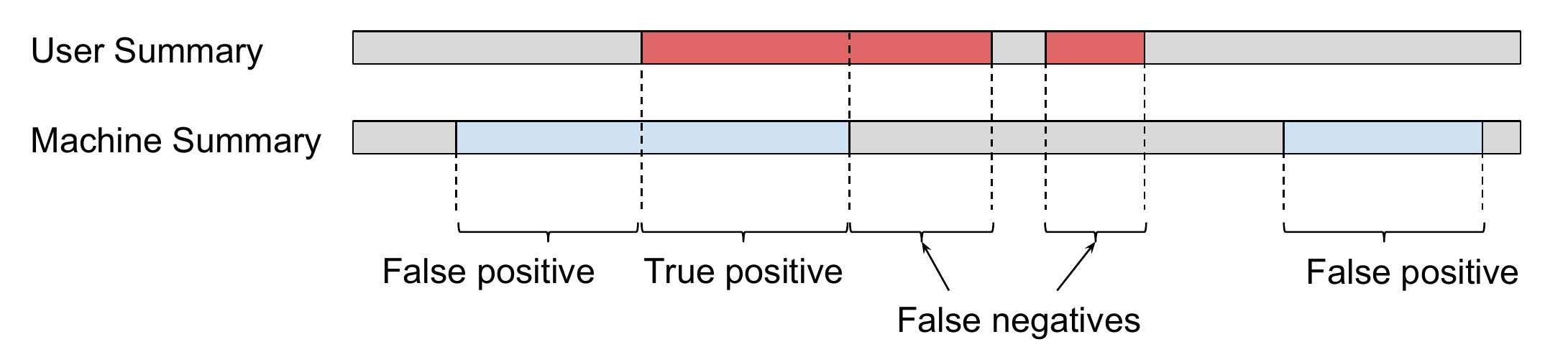}
	\end{center}
	\caption{True positives, False positives and False negatives are calculated per-frame between the ground truth and machine binary keyshot summaries.}
	\label{fig:prec_rec}
\end{figure}

To provide a fair comparison with the state of the art, we follow evaluation protocol from 
\cite{zhang2016video}, \cite{Zhou2018DeepRL} and \cite{Mahasseni2017UnsupervisedVS}.
To asses the similarity between the machine and user summaries we use the harmonic mean of precision and recall expressed as the F-score in percentages.
\begin{equation} \label{eq:f_score}
F=2\times {\frac {\mathrm {precision} \times \mathrm {recall} }{\mathrm {precision} +\mathrm {recall} }} \times 100
\end{equation}
True and false positives and false negatives for the F-score are calculated per-frame as the overlap between the ground truth and machine summaries, as shown in Fig. \ref{fig:prec_rec}.

Following \cite{GygliECCV14}, the machine summary is limited to 15\% of the original video length
and then evaluated against multiple user summaries according to \cite{zhang2016video}. Precisely, on the TvSum benchmark, for each video, the F-score is calculated as an average between the machine summary and each of the user summaries as suggested by \cite{song2015tvsum}. Average F-score over videos in the dataset is then reported. On the SumMe benchmark, for each video, a user summary most similar to the machine summary is selected. This approach is proposed by \cite{gygli2015video} and also used in the work of Lin and Chin-Yew \cite{lin2004rouge}.

\section{Experiments and Results}

Results of the VASNet evaluation on TvSum and SumMe datasets, compared with the most recent state of the art methods are presented in Table \ref{table:sota_table}. 
To illustrate how well the methods learned from the user annotations we show a human performance, which is calculated as pairwise F-scores between the ground truth and all user summaries.
In Table \ref{table:datasets_fscores} we also compare the human performance with F-scores calculated among the user summaries themselves. 
\setlength{\tabcolsep}{4pt}
\begin{table}
	\begin{center}
		\caption{Average pairwise F-scores calculated among user summaries and between ground truth (GT) and users summaries.}		
		\label{table:datasets_fscores}
		\begin{tabular}{l|c|c}
			\hline
			& \multicolumn{2}{l}{\textbf{Pairwise F score}}                                                                                                                        
			\\ 
			\cline{2-3}
			\textbf{Dataset} & \textbf{\begin{tabular}[c]{@{}l@{}}Among users \\ annotations\end{tabular}} & \textbf{\begin{tabular}[c]{@{}l@{}}Training GT w.r.t.\\ users annotations\\(human performance)\end{tabular}} \\ 
			\hline
			\textbf{SumMe} & 31.1 & 64.2\\ 
			\hline
			\textbf{TvSum} &53.8 &63.7 \\ 
			\hline
		\end{tabular}
	\end{center}
\end{table}
We can see that the human performance is higher than the F-score among the user summaries which is likely caused by the fact that the training ground truth is calculated as an average of all user summaries and then converted to the keyshots, which are aligned on the scene change-points. These keyshots are likely to be longer than the discrete user summaries, thus having higher mutual overlap. The pairwise F-score $53.8$ for TvSum dataset is higher than the F-score $36$ reported by the authors \cite{song2015tvsum}. This is because we convert each user summary to keyshots with KTS and limit the duration to 15\% of the video length and then calculate the pairwise F-scores. Authors of the dataset \cite{song2015tvsum} calculate the F-score from \textit{gold standard labels}, that is, from keyshots of length 2 seconds, a length used by users during the frame-level score annotation. We chose to follow the former procedure which is maintained in all evaluations in this work to make the results directly comparable.

\begin{figure}%
	\centering
	\subfloat[SumMe dataset]{{\includegraphics[width=5.82cm]{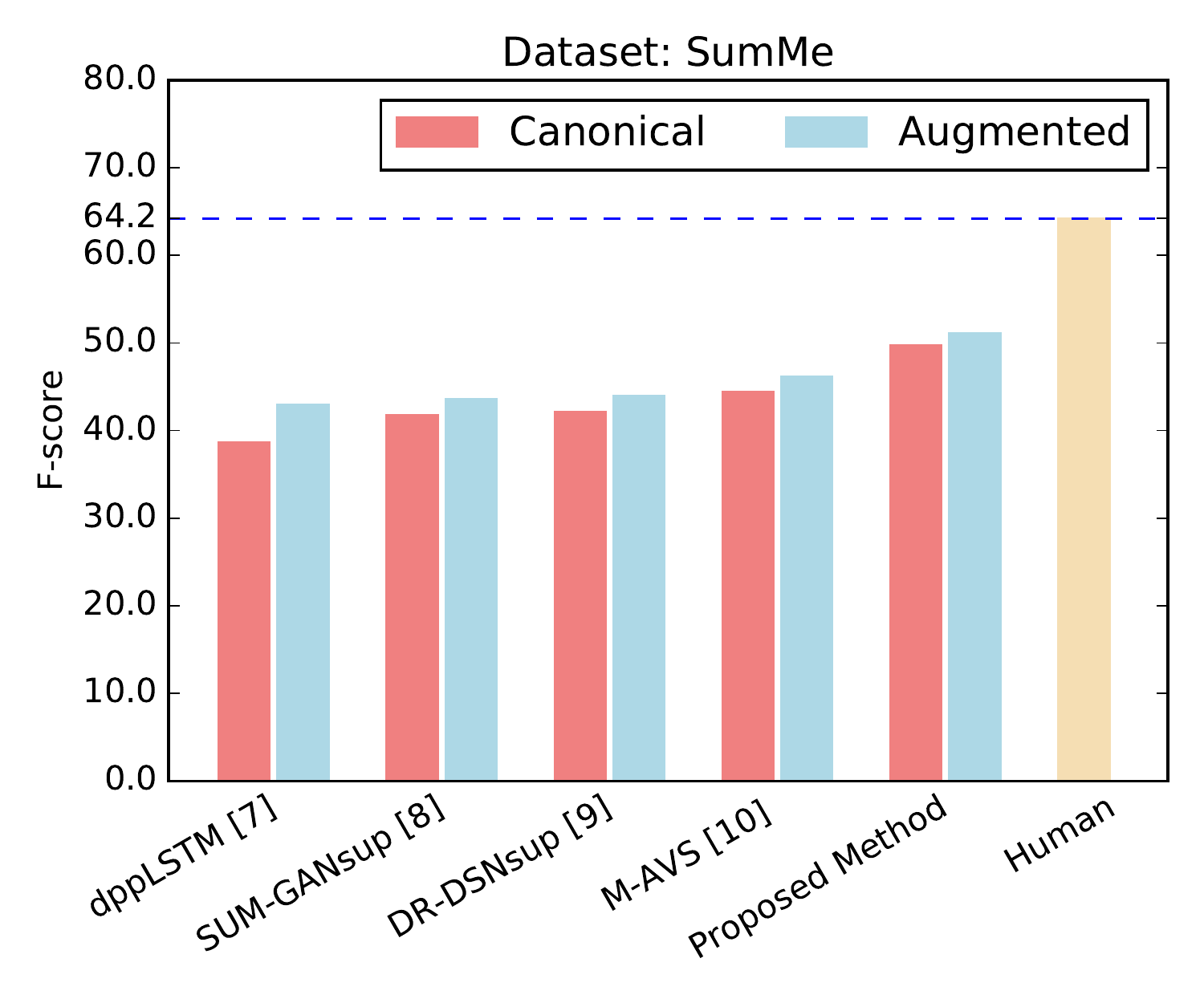} }}
	\quad
	\subfloat[TvSum dataset]{{\includegraphics[width=5.82cm]{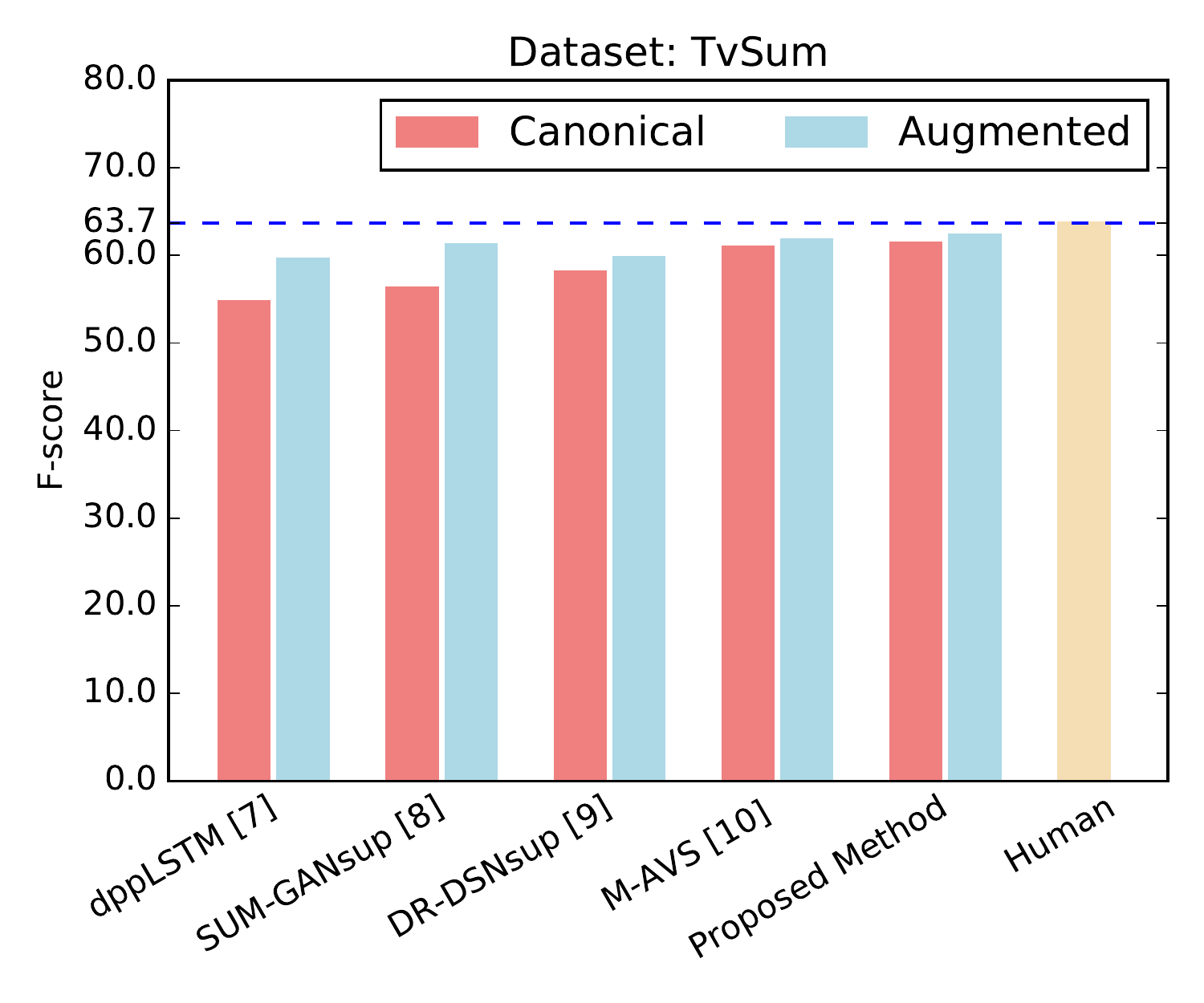} }}
	\caption{VASNet performance gain compared to the state of the art and human performance.}
	\label{fig:sota_chars}
\end{figure}

\setlength{\tabcolsep}{4pt}
\begin{table}
	\begin{center}
		\caption{Comparison of our method VASNet with the state of the art methods for canonical and augmented settings. For a reference we add human performance measured as pairwise F-score between training ground truth and user summaries.}
		\label{table:sota_table}
		\begin{tabular}{c||c|c||c|c}
			\hline
			& \multicolumn{2}{c||}{\textbf{SumMe}}  & \multicolumn{2}{c}{\textbf{TvSum}}    \\ 
			\cline{2-5}
			\textbf{Method} & \textbf{Canonical} & \textbf{Augmented}  & \textbf{Canonical} & \textbf{Augmented} \\ 
			\hline 
			
			\hline
			
			dppLSTM \cite{zhang2016video} & 38.6 & 42.9  & 54.7 & 59.6 \\ 
			\hline
			M-AVS \cite{ji2017video} & 44.4 & 46.1 & 61.0 & 61.8 \\ 
			\hline
			$\text{DR-DSN}_{sup}$ \cite{Zhou2018DeepRL}&42.1  & 43.9  & 58.1 & 59.8 \\ 
			\hline
			$\text{SUM-GAN}_{sup}$ \cite{Mahasseni2017UnsupervisedVS} & 41.7 & 43.6 & 56.3&61.2\\ 
			\hline
			$\text{SASUM}_{sup}$ \cite{DBLP:conf/aaai/WeiNYYYY18} & 45.3 & - &58.2& - \\ 
			\hline
			Human  & 64.2 & - & 63.7 & - \\
			\hline
			\begin{tabular}[c]{@{}c@{}}VASNet\\(proposed method)\end{tabular}  & \textbf{49.71} & \textbf{51.09} & \textbf{61.42} & \textbf{62.37} \\
			\hline
		\end{tabular}
	\end{center}
\end{table}
\setlength{\tabcolsep}{1.4pt}
In Table \ref{table:sota_table} we can see that our method outperforms all previous work in both canonical and augmented settings. On the TvSum benchmark the improvement is by 0.7\% and 1\% in the canonical and augmented settings respectively and 2\% lower than the human performance. On the SumMe this is 12\% and 11\% in the canonical and augmented settings respectively and 21\% below the human performance. In Fig. \ref{fig:sota_chars} we show this improvements visually.

The higher performance gain on the SumMe dataset is very likely caused by the fact that our attention model can extract more information from the ground truth compared to the TvSum, where most methods already closely approach the human performance. It is conceivable to assume that the small gain on the TvSum is caused by the negative effect of the global attention on long sequences. 
TvSum videos are comparatively longer than the SumMe as seen in Table \ref{table:datasets_properties}. 
At every prediction step the global attention 'looks' at all video frames. 
For long video sequences frames from temporally distant scenes are likely less relevant than the local ones, but the global attention still needs to explore them. 
We believe that this increases variance in the attention weights, which negatively impacts the prediction accuracy. We hypothesize that this could be mitigated by the introduction of local attention. 

\subsection{Qualitative Results}
\begin{figure}[h]
	\centering
	\subfloat{\includegraphics[width=0.52\linewidth]{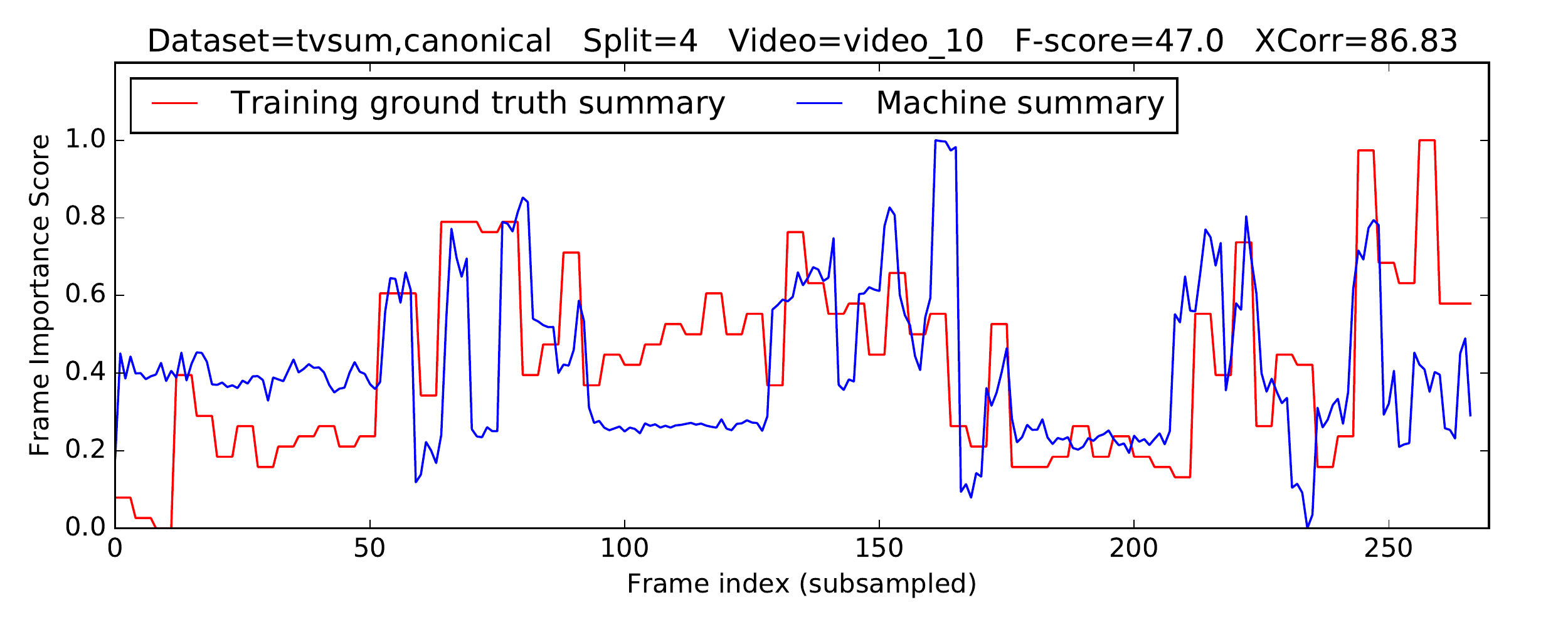}}
	\subfloat{\includegraphics[width=0.52\linewidth]{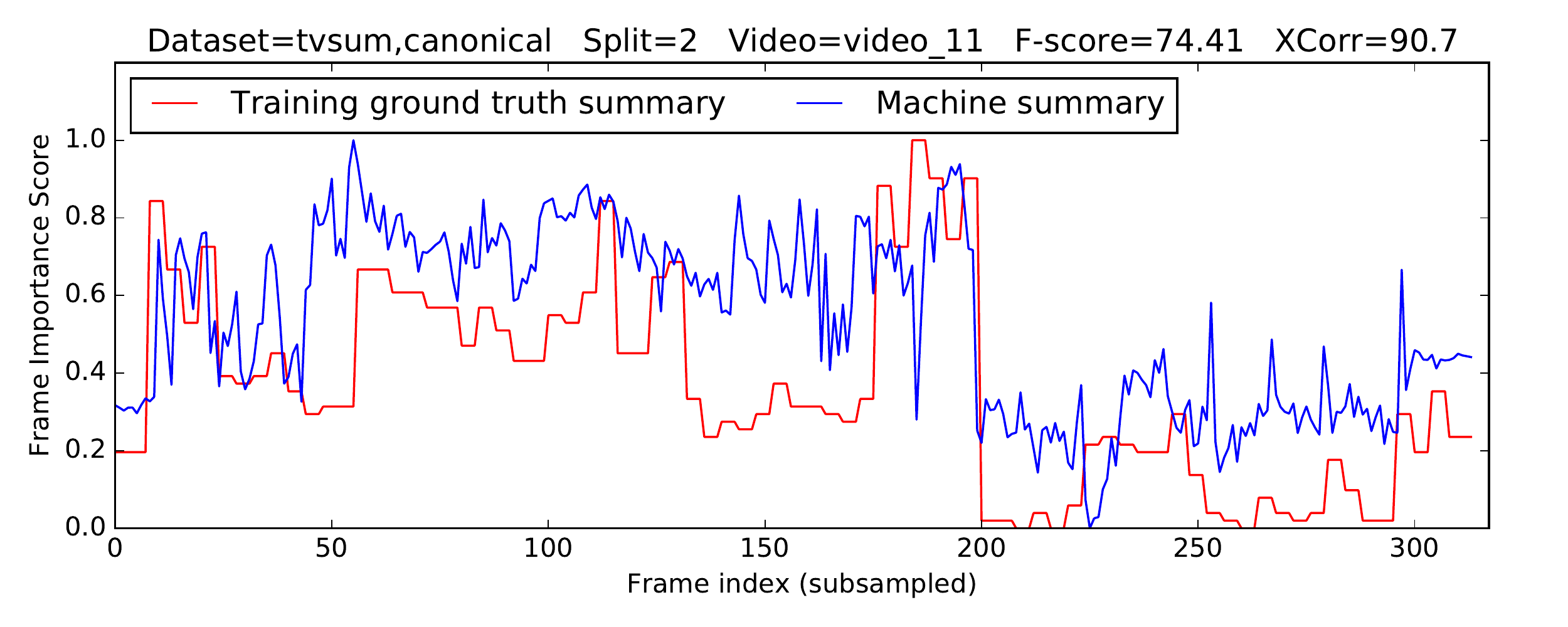}}
	\caption{Correlation between ground truth and machine summaries produced by VASNet for test videos 10 and 11 from TvSum dataset, also evaluated in \cite{Zhou2018DeepRL}. }
	\label{fig:corr_tvsum} 
\end{figure}

To show the quality of the machine summaries produced by our method we plot the ground truth and predicted scores for two videos from TvSum  in Fig. \ref{fig:corr_tvsum}.
We selected videos 10 and 11, since they are also used in previous work \cite{Zhou2018DeepRL}, 
thus enabling a direct comparison. We can see a clear correlation between the ground truth and machine summary, confirming the quality of our method. Original videos and their summaries are available on YouTube. \footnote{\url{https://www.youtube.com/playlist?list=PLEdpjt8KmmQMfQEat4HvuIxORwiO9q9DB}}

\begin{figure}[ht]
	\begin{center}
		\vspace{-10pt}
		\includegraphics[width=0.8\linewidth]{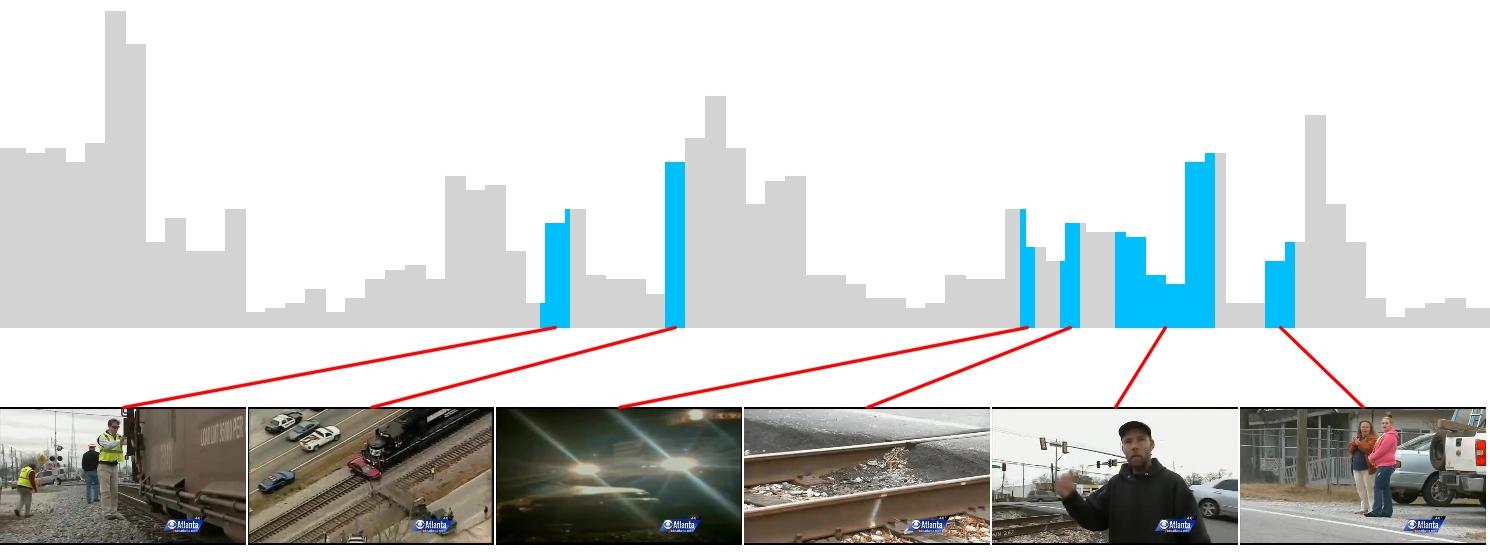}
		\caption{Ground truth frame scores (gray), machine summary (blue) and corresponding  keyframes for test video 7 from TvSum dataset. 
		}
		\label{fig:keyshots_stripe}
	\end{center}
\end{figure}

\begin{figure}[ht]
	\begin{center}
		\vspace{-10pt}
		\includegraphics[width=1.0\linewidth]{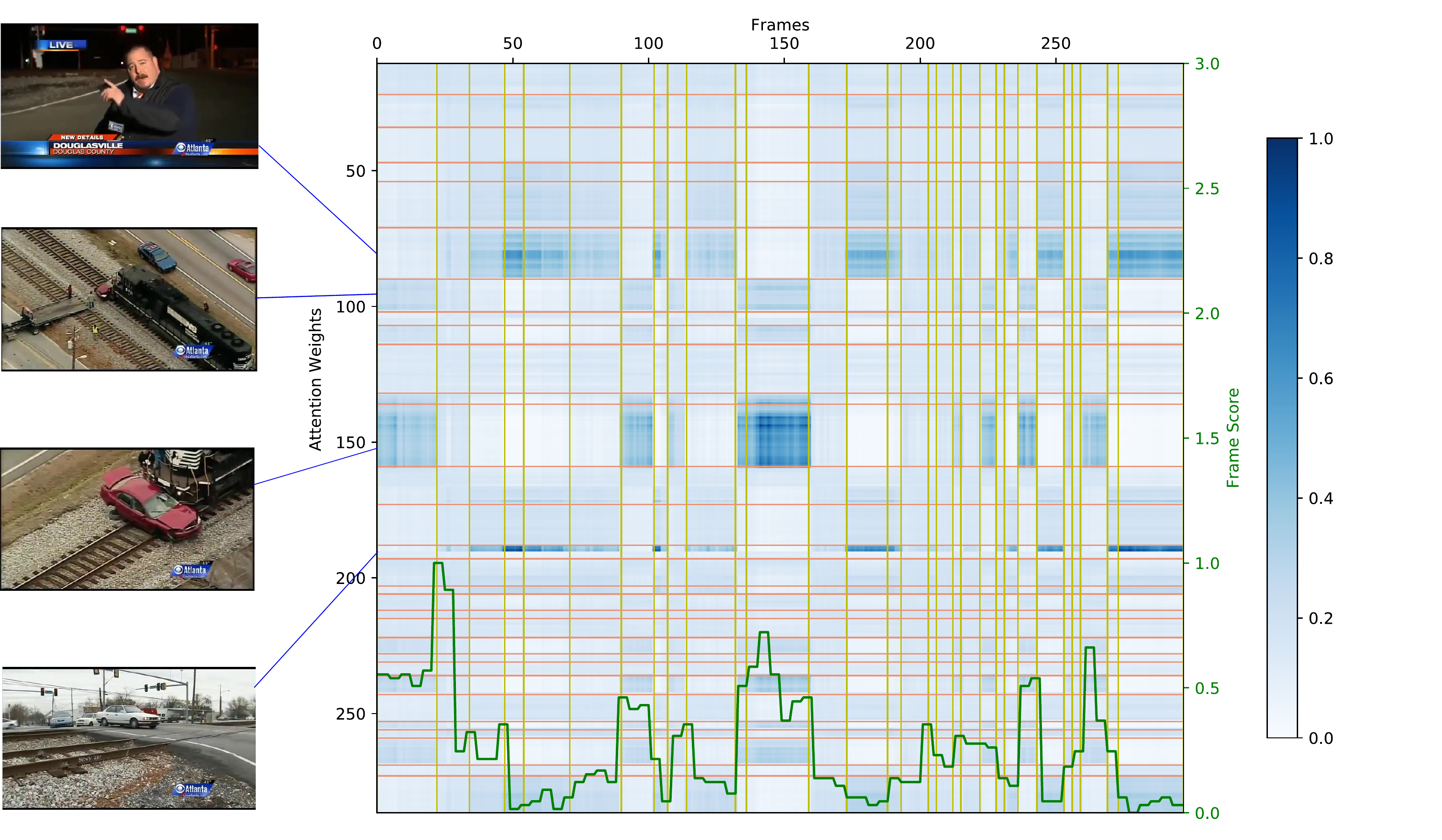}
		\caption{Confusion matrix of attention weights for TvSum video 7 from test split 2. 
			Green plot at the bottom shows the GT frame scores. Green and red horizontal and vertical lines show scene change points. Values were normalized to range 0-1 across the matrix. Frames are sub-sampled to 2fps.}
		\label{fig:conf_mat}
		\vspace{-10pt}
	\end{center}
\end{figure}

We also compare the final, binary keyshot summary with the ground truth. In Fig. \ref{fig:keyshots_stripe} we show machine generated keyshots in light blue color over the ground truth importance scores shown in gray. We can see that the selected keyshots align with most of the peaks in the ground truth and that they cover the entire length of the video.
The confusion matrix in Fig. \ref{fig:conf_mat} shows attention weights produced during evaluation of TvSum video 7. 
We can see that the attention strongly focuses on frames either correlated with low frame scores (top and bottom image in Fig. \ref{fig:conf_mat}, attention weights for frames $\sim$80 and $\sim$190)  or high scores (second and third image, frames $\sim$95 and $\sim$150). It is conceivable to assume that the network learns to associate every video frame with other frames of similar score levels. 

Another interesting observation to make is that the transitions between the high and low attention weights in the confusion matrix highly correlate with the scene change points, shown as green and red horizontal and vertical lines. It is important to note that the change points, detected with KTS algorithm, were not provided to the model during learning or inference, nor were used to process the training GT. Thus, we believe that this model could be also applied to  scene segmentation, removing the need for the KTS post-processing step. We will explore this possibility in our future work.

\section{Conclusions}
In this work propose a novel deep neural network for keyshot video summarization based on pure soft, self-attention. This network performs a sequence to sequence transformation without recurrent networks such as LSTM based encoder-decoder models. We show that on the supervised, keyhost video summarization task our model outperforms the existing state of the art methods on the TvSum and SumMe benchmarks. Given the simplicity of our model it is easier to implement and less resource demanding to run than LSTM encoder-decoder based methods, making it suitable for application on embedded or low power platforms.

Our model is based on a single, global, self-attention layer followed by two, fully connected network layers. We intentionally designed and tested the simplest architecture with global attention, and without positional encoding to establish a baseline method for such architectures. Limiting the aperture of the attention to a local region as well as adding the positional encoding are simple modifications that are likely to further improve the performance. We are considering these extensions for our future work.

The complete PyTorch 0.4 source code to train and evaluate our model, as well as trained weights to reproduce results in this paper, will be publicly available on \url{https://github.com/ok1zjf/VASNet}.


\clearpage
\bibliographystyle{splncs04}
\bibliography{ok1zjf-v4}

\end{document}